\documentclass[11pt,a4paper]{article}
\usepackage[hyperref]{emnlp2020}
\usepackage{times}
\usepackage{latexsym}

\usepackage{url}

\aclfinalcopy % Uncomment this line for the final submission
\usepackage{amsmath}
\usepackage{multirow}
\usepackage{amssymb}
\usepackage{mathrsfs}
\usepackage{graphics}
\usepackage{graphicx}
\usepackage{ifsym}
\makeatletter
\g@addto@macro\normalsize{
	\setlength\abovedisplayskip{1pt}
	\setlength\belowdisplayskip{1pt}
	\setlength\abovedisplayshortskip{1pt}
	\setlength\belowdisplayshortskip{1pt}
}
\makeatother

\usepackage{subcaption}
\usepackage[rightcaption]{sidecap}
\usepackage{booktabs}
\usepackage{multirow}
\usepackage{array}
\usepackage{bbm}
\usepackage{multicol}
\usepackage{blindtext}
\usepackage[labelfont={small,bf},font={small,bf},skip=3pt]{caption}

\usepackage{algorithm}
\usepackage{algpseudocode}
\usepackage{tikz}
\usepackage{makecell}
\usepackage{xcolor,colortbl}
\usepackage{threeparttable}

\usepackage{array}
\newcolumntype{H}{>{\setbox0=\hbox\bgroup}c<{\egroup}@{}}

\newcommand{\vect}[1]{\boldsymbol{\mathrm{#1}}}

\newcommand{\unc}{\boldsymbol{\omega}}
\newcommand{\con}{\boldsymbol{\mu}}
\newcommand{\dir}{\boldsymbol{\alpha}}

\newcommand{\expect}[1]{\ensuremath{\underset{#1}{\mathbb{E}}\xspace}}

\newenvironment{packed_item}{
	\begin{itemize}
		\setlength{\itemsep}{0pt}
		\setlength{\parskip}{0pt}
		\setlength{\parsep}{0pt}
	}
	{\end{itemize}}

\newenvironment{packed_enum}{
	\begin{enumerate}
		\setlength{\itemsep}{0pt}
		\setlength{\parskip}{0pt}
		\setlength{\parsep}{0pt}
	}
	{\end{enumerate}}

\graphicspath {{figure/}}

\title{Modeling Token-level Uncertainty to Learn Unknown Concepts in SLU \\ via Calibrated Dirichlet Prior RNN}

\author{
	Yilin Shen\textsuperscript{1}, Wenhu Chen\textsuperscript{2}, Hongxia Jin\textsuperscript{1} \\
	\textsuperscript{1}Samsung Research America, Mountain View, CA, USA \\
	\textsuperscript{2}University of California, Santa Barbara, CA, USA \\
	{\tt \footnotesize \{yilin.shen,hongxia.jin\}@samsung.com, wenhuchen@cs.ucsb.edu}
}

\begin{document}

\maketitle

\begin{abstract}
	One major task of spoken language understanding (SLU) in modern personal assistants is to extract semantic concepts from an utterance, called slot filling.
	Although existing slot filling models attempted to improve extracting new concepts that are not seen in training data, the performance in practice is still not satisfied.
% 	Existing slot filling models are trained on pre-collected data and suffer from the failure to extract unseen concepts.
	Recent research collected question and answer annotated data to learn what is unknown and should be asked, yet not practically scalable due to the heavy data collection effort.
	In this paper, we incorporate softmax-based slot filling neural architectures to model the sequence uncertainty without question supervision.
	We design a Dirichlet Prior RNN to model high-order uncertainty by degenerating as softmax layer for RNN model training.
	To further enhance the uncertainty modeling robustness, we propose a novel multi-task training to calibrate the Dirichlet concentration parameters.
	We collect unseen concepts to create two test datasets from SLU benchmark datasets Snips and ATIS.
	On these two and another existing Concept Learning benchmark datasets, we show that our approach significantly outperforms state-of-the-art approaches by up to 8.18\%.
	Our method is generic and can be applied to any RNN or Transformer based slot filling models with softmax layer.
\end{abstract}

\section{Introduction}

With the rise of modern artificially intelligent voice-enabled personal assistants (PA) such as Alexa, Google Assistant, Siri, etc., spoken language understanding (SLU) plays a vital role to understand all varieties of user utterances and carry out the intent of users.
One of the major tasks in SLU is \emph{slot filling}, which aims to extract semantic values of predefined slot types from a natural language utterance.
Existing slot filling approaches are only designed for offline model training based on a large scale pre-collected training corpus. 
They explored to use machine learning and deep learning techniques for an independent slot filling model \cite{kurata2016leveraging,hakkani2016multi,liu2016attention} or a joint learning model with intent detection
 \cite{liu2016attention,Yu2018bimodel,Goo2018slotgated}.

\begin{figure}[t]
	\centering
	\includegraphics[width=1\columnwidth]{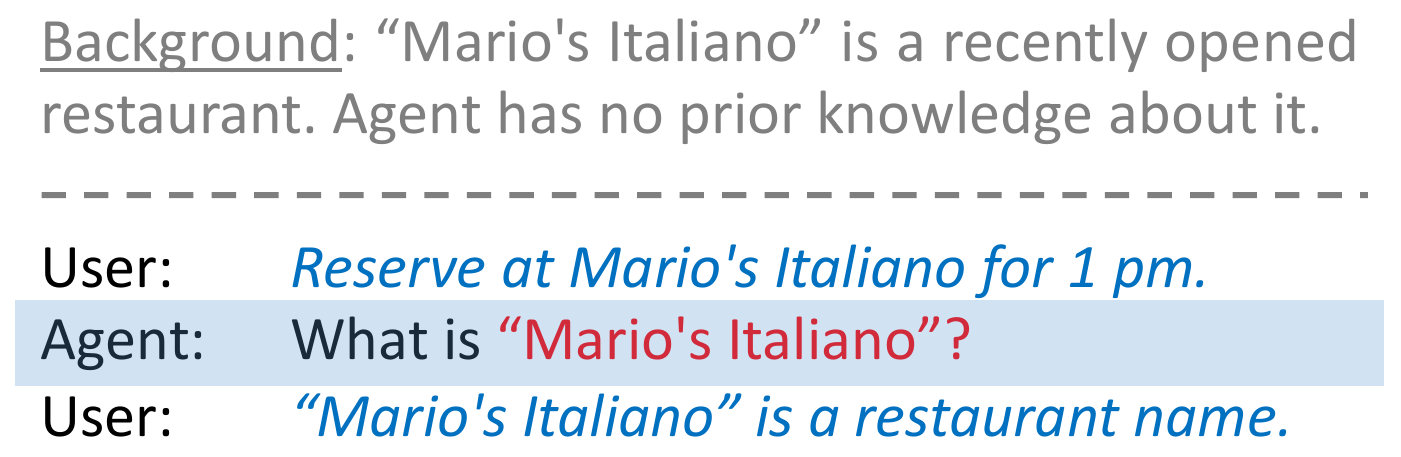}
	\caption{Example of Asking Clarification Question}
	\label{fig:example}
\end{figure}

However, real usage data are typically very different from pre-collected training data.
Many slot types can have a large number of new slot values, e.g., book, music and movie names.
For example, the benchmark Snips dataset only covers less than 0.01\% of the existing slot values, let alone their fast growth.
Also, a user could have his personalized slot values for specific meanings, e.g., ``happy hours" as a playlist name.
In this paper, we define \emph{concept} as new slot values for a slot type (while a new slot type can be supported similarly).
While some new concepts could be extracted using emerging models \cite{lin2017emerging,jansson2017emerging} or recently improved slot filling models to tackle out-of-vocabulary (OOV) using character embedding \cite{liang2017character}, copy mechanism \cite{Zhao18pointer} and few-shot learning \cite{Hu19oov,hou2019fewshot,Fritzler2019fewshot}, these models still cannot extract a majority of new concepts, called \emph{unknown concepts}.
Thus, it is critically desirable to enable PA to smartly detect and clarify the unknown concept (as in~\autoref{fig:example}) in order to take the correct action.

Earlier work explored rule based approaches to ask clarification questions~\cite{Hori2003ODQA,Purver06clarie}.
However, these approaches are not applicable to modern deep learning SLU models.
Recently, research communities raised special interest on how to make machine learning models know what it does not know~\cite{rajpurkar2018squad} and how to ask clarification questions~\cite{rao2018clarification}.
% \footnote{Both papers are selected as best papers in ACL 2018.}.
However, these dataset and approach are designed based on the assumption that question/answer supervision is provided.
Another line of work studied uncertainty estimation on classification problems~\cite{hendrycks2016baseline,liang2017enhancing,chen2018variational}, yet not applicable to RNN based models.
Unfortunately, due to the large amount and fast growth of new concepts, it is not scalable to continuously collect and annotate new training data.
The most relevant work~\cite{jia2017concepts} proposed a hybrid approach to combine model confidence, dependency tree and out-of-vocabulary words to derive the uncertainty of each word in an utterance.
~\cite{li2018confidence} further proposed a few metrics to measure confidence. 
However, its performance suffers from the over-confidence of unseen concepts in pre-trained slot filling models.

% \todo{OOD data is infinite..}
% \todo{Existing method used pointer network, copy mechansim to mitigate this problem.. however not sufficient}

In this paper, we focus on modeling the token-level uncertainty to extract unknown concepts \emph{without unknown concept supervision}.
That said, our model is only trained on existing concepts in training data.
Due to the infinite number and fast growth of new concepts in reality, our setting is more practically reasonable.
Our approach takes any softmax-based slot filling model and outputs the uncertainty score of each word.
Inspired by Dirichlet prior network~\cite{malinin2018prior}, we first design a Dirichlet Prior RNN by degenerating Dirichlet prior into the softmax layer in slot filling models.
Next, we design a multi-task algorithm to further calibrate the Dirichlet concentration parameters to enhance the robustness of uncertainty modeling by learning an adaptive calibration matrix.
At last, we incorporate the utterance syntax with words with high uncertainty to extract the unknown concept phrases.
Our contributions can be summarized as follows:
\begin{packed_item}
	\item We design Dirichlet prior RNN to model high-order token-level uncertainty by degenerating softmax-based slot filling models;
	\item We propose a multitask learning algorithm to calibrate the Dirichlet concentration parameters to improve uncertainty estimation;
% 	\item We collect unseen concepts to create test sets on benchmark Snips and ATIS datasets;
	\item Our approach achieves state-of-the-art performance on three SLU benchmark datasets;
	\item Our method is generic which can also been applied onto Transformer/BERT based slot filling models with softmax output layers.
\end{packed_item}

%http://mentalfloss.com/article/85305/how-many-books-have-ever-been-published
%https://thewebminer.com/blog/how-many-restaurants-are-in-the-world
%https://www.imdb.com/pressroom/stats/
%https://stephenfollows.com/how-many-films-are-released-each-year/

\section{Related Work}

% \begin{figure*}[t]
% 	\centering
% 	\includegraphics[width=0.8\textwidth]{dirichlet}
% 	\caption{Illustration of uncertainty on the simplex $\mathbb{S}_k$ in Dirichlet prior networks}
% 	\label{fig:dirichlet}
% \end{figure*}

\noindent\textbf{Slot Filling:}
Slot filling task is to label each word in an utterance to one of the predefined slot tags.
\cite{xu2013convolutional} first proposed to use the convolutional neural network (CNN) together with a conditional random field (CRF).
Recently, many researches have surged based on recurrent neural networks (RNN) models.
\cite{liu2015recurrent,mesnil2015using,peng2015recurrent} first uses RNN in a straight-forward way to generate multiple semantic tags sequentially by reading in each word one by one.
Later on, \cite{liu2016attention} presented an attention mechanism by incorporating the utterance context.
%It also introduced a joint model to learn slot filling task along with intent classification, the other main task in NLU.
\cite{Yu2018bimodel} designed a bi-model via asynchronous training.
\cite{Goo2018slotgated} proposed to learn the relationship between intent and slot attention vectors.
However, these approaches are only designed to extract IND concepts.

%\todo{RL based approach...}

\noindent\textbf{Learning to Clarify:}
Earlier researches apply syntax rules to detect unknown phrases~\cite{Schlangen01resolvingunderspecification,Hori2003ODQA,Purver06clarie}.
However, we have shown in experiments that these approaches are insufficient to achieve good performance.
%Another line of work is asking clarification question at utterance level.
%CLARIE~\cite{Purver06clarie} applied rules to detect if the agent can understand the entire user utterance, yet it cannot extract the unknown concept from an utterance.
~\cite{li2018confidence} proposed statistical metrics to measure deep learning based semantic parser.
Yet, these approaches cannot extract unseens concepts within one utterance.
Recently, ~\cite{rajpurkar2018squad} collected a QA dataset with question/answer supervision to encourage the research community to model an agent which knows what it does not know.
~\cite{rao2018clarification} proposed a supervised question ranking approach to select the most relevant clarification questions.
However, such supervised approach is not scalable in practice due to heavy data collection workload.

\noindent\textbf{Uncertainty Estimation:}
The most relevant work~\cite{jia2017concepts} designed a simple approach to combine slot filling model confidence score, utterance syntax and OOV words to estimate the uncertainty score and extract unseen concepts.
\cite{li2018confidence} studies different uncertainty metrics for semantic parser models.
Many recent researches in computer vision community make great progress in uncertainty model in CNN, including a baseline~\cite{hendrycks2016baseline}, temperature scaling approaches~\citep{liang2017enhancing,lee2017training,shalev2018out,devries2018learning}, adversarial training~\citep{lee2017training} and variational inference ~\cite{chen2018variational,malinin2018prior}.

\section{Background \& Problem Definition}\label{sc:background}

\setlength{\tabcolsep}{0.1em}
\begin{table}[t]\scriptsize
	\centering
	\caption{Uncertainty Modeling Example for Slot filling (In label uncertainty, L=low and H=high)}
	\label{table:atis_example}
	\begin{tabular}{c|ccccccc}
		\toprule
		\textbf{Utterance} & reserve & at & Mario's & Italiano & for & 1 & pm \\
		\midrule
		\midrule
		\textbf{True Labels} & O & O & B-restaurant & B-restaurant & O & B-time & I-time \\
		\midrule
		\textbf{Predicted Labels} & O & O & \color{red}{B-game} & \color{red}{B-country} & O & B-time & I-time \\
		\textbf{Label Uncertainty} & L & L & H & H & L & L & L \\
		\textbf{Final predicted Labels} & O & O & \textbf{B-unknown} & \textbf{I-unknown} & O & B-time & I-time \\
		\bottomrule
	\end{tabular}
\end{table}

\subsection{Slot Filling}

Slot filling is to extract semantic concepts from a natural language utterance.
It has been modeled as a sequence labeling problem in literature.
Given an input utterance as a sequence of words $\vect{x} = (x_1, \ldots, x_n)$ of length $n$, slot filling maps $\vect{x}$ to the corresponding label sequence $\vect{y}$ of the same length $n$.
Each word is labeled as one of the total $K$ labels.
As shown in~\autoref{table:atis_example}, each output label in $\vect{y}$ is in the format of IOB, where ``B" and ``I" stand for the beginning and intermediate word of a slot and ``O" means the word does not belong to any slot.
As a sequence labeling problem, slot filling can be solved using traditional machine learning approaches \cite{McCallum2000MEM,Raymond2007GenerativeAD} and recent mainstream recurrent neural network (RNN) based approaches which takes and tags each word in an utterance one by one \cite{Yao2014slottagging,mesnil2015using,peng2015recurrent,liu2015recurrent,kurata2016leveraging}.
These RNN based approaches designed different models to estimate the maximum likelihood:
\begin{equation*}
P(\theta|D)= \max_{\theta} \displaystyle E_{D} \Big[ \prod_{t=1}^n P(y_t | y_1, \ldots, y_{t-1}, \vect{x}; \theta) \Big]
\end{equation*}
By optimizing the following loss function:
\begin{equation}\label{eq:rnn_obj}
\mathcal{L} (\theta) \triangleq - {1 \over n} \sum_{i=1}^{|S|} \sum_{t=1}^n y_t(i) \log P_t(i)
\end{equation}
where $D= \{\vect{x}, \vect{y}\} \sim p(\vect{X}, \vect{Y})$ is the training set;
$\theta$ are the model parameters.
$P_t(i)$ is typically computed using softmax function.
All models with such loss function are referred to as \emph{softmax RNN slot filling model}.

\subsection{Problem Definition}

We refer to the training data $D$ as in-distribution (IND) data and each utterance in $D$ is drawn from a fixed but unknown distribution $P_{IND}$.
Our goal is to train a slot filling model \emph{only} on IND data $D$ to achieve two objectives:
(1) correctly label a test sequence $\vect{z}$ drawn from the same distribution $P_{IND}$;
(2) identify out-of-distribution elements in a test sequence $\vect{z}$ drawn from a different distribution $P_{OOD}$, referred to as out-of-distribution (OOD) data.
Note that the OOD sequence may only have a subset of out-of-distribution elements.
In this case, our goal is to identify only these OOD elements and correctly label other IND elements.

Specifically, we take a pre-defined softmax RNN slot filling model architecture (described in previous subsection) as input.
We train our calibrated prior RNN model only by enhancing its softmax layer.
Our model outputs two sequences for each test sequence $\vect{z}$ with an unseen concept $\vect{c}^*$:
(1) traditional label sequence $\vect{y}$;
(2) uncertainty sequence $\vect{u}$ with an uncertainty score $u_t$ for each word $z_t$.
For $\vect{z}$ drawn from $P_{IND}$, all labels in $\vect{y}$ are expected to be correct and all uncertainty scores in $\vect{u}$ are expected to be low.
For $\vect{z}$ drawn from $P_{OOD}$, the model is expected to either label an element correctly or gives it a high uncertainty score.
At last, the model tags all concepts with correct slot types or as unknown.

\section{Calibrated Dirichlet Prior RNN}

\begin{figure}[t]
	\centering
	\includegraphics[width=1.0\columnwidth]{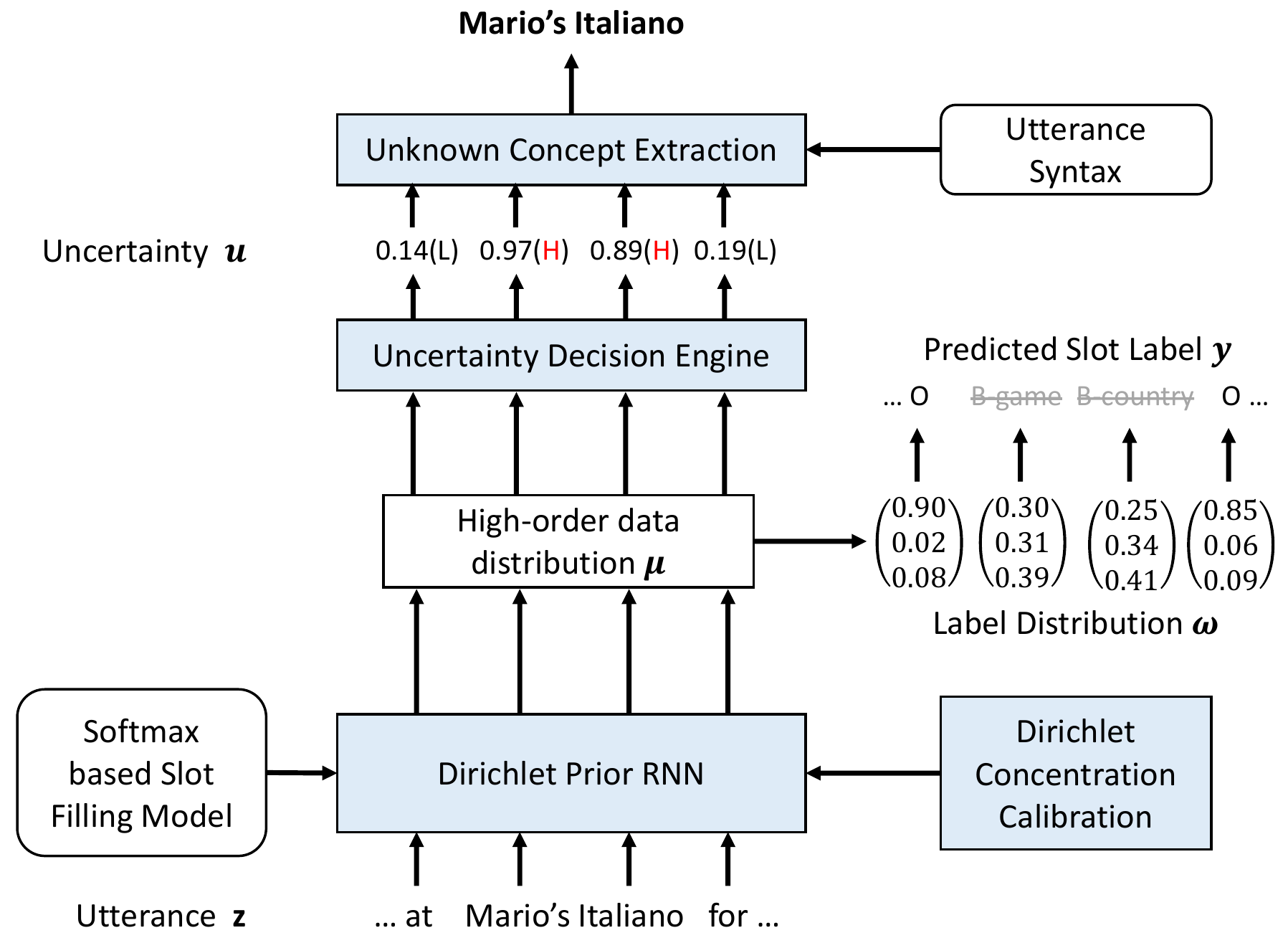}
	\caption{Method Overview:
		Dirichlet prior RNN is the main component in our method that is degenerated as traditional softmax based training on given softmax slot filling models.
		The calibration module introduces an additional objective as a multi-task training to better separate the IND and OOD concepts.
% 		The enhanced model is designed as a multi-task training that optimizes both traditional RNN softmax and the calibration objectives simultaneously.
		During testing, we output the label distribution $\unc_t$ for $t^{th}$ word in utterance $\vect{z}$.
		Based on $\unc_t$, the predicted label $y_t$ and uncertainty estimation $u_t$ is computed.
		When $u_t$ is larger than some threshold, we set its uncertainty label to H and consider the predicted label $y_t$ as unknown.
		At last, we extract the unknown concept by expanding the words with high uncertainty score and utterance's dependency tree.
	}
	\label{fig:model}
\end{figure}

%This problem is further deteriorated in our sequence modeling problem.
~\autoref{fig:model} shows the overview of our proposed method that is designed to incorporate with any slot filling models with softmax output layer.
In this section, we focus on presenting two main novel components in blue.

% \todo{RNN architecture can include word/character embedding etc.}

\subsection{Dirichlet Prior RNN}

As discussed in Section \ref{sc:background}, the existing approaches for slot filling are discriminative models which aim to find the best model parameters $\theta^*$ to fit training data.
To estimate the uncertainty of each word, a naive solution is to use its softmax confidence at each step.
However, since an unseen concept is out of distribution from the training data, such pretrained neural networks tend to be blindly overconfident on predictions of completely unrecognizable \cite{Nguyen2015fool} or irrelevant inputs \cite{hendrycks17baseline,Moosavi2017uap}.

Consider the IND training set $D = \{\vect{x}, \vect{y}\} \sim p(\vect{X}, \vect{Y})$.
For an utterance $\vect{z}$, we model its high-order distribution $p(\unc|\vect{z}, D)$ using a Bayesian framework as follows:
\begin{align}
p(\unc |\vect{z}, D) = \int \underbrace{p(\unc| \vect{z}, \theta)}_\text{Data} \underbrace{p(\theta|D)}_\text{Model} d\theta
\label{eq:ensemble}
\end{align}
where, similar as in~\autoref{eq:rnn_obj}, we have:
$$p(\unc| \vect{z}, \theta) = \prod_{t=1}^n p(\unc_t | \unc_1, \ldots, \unc_{t-1}, \vect{z}; \theta)$$
where $\unc_i$ is the predicted distribution of all possible $K$ labels for each word $i$ in $\vect{z}$.
The \emph{data uncertainty} is described by label-level posterior $p(\unc|\vect{z}, \theta)$ and \emph{model uncertainty} is described by model-level posterior $p(\theta|D)$.
However, the integral in~\autoref{eq:ensemble} is intractable in deep neural networks, thus Monte-Carlo Sampling algorithm is used to approximate it as follows:
\begin{eqnarray*}
	p(\unc| \vect{z}, D) \approx \frac{1}{M} \sum_{i=1}^M p(\unc| \vect{z}, \theta^{(i)});
	\theta^{(i)} \sim p(\theta|D)
\end{eqnarray*}
where each $p(\unc_t | \unc_1, \ldots, \unc_{t-1}, \vect{z}, \theta^{(i)})$ is a categorical distribution $\con$ over the simplex with $\con=[\mu_1, \cdots, \mu_K] = [p(y=\omega(1), \cdots, p(y=\omega(K))]^T$.
This ensemble is a collection of points on the a simplex which can be viewed as an implicit distribution induced by the posterior over the model parameters $p(\theta|D)$.

In addition to the intractability of posteriori estimation, such ensemble of implicit distribution also leads to the challenge of measuring uncertainty.
For example, a high entropy of $p(\unc| \vect{z}, D)$ could indicate uncertainty in the prediction due to either an IND input in a region of class overlap or an OOD input far from the training data.
When using other measures like mutual information~\cite{gal2016uncertainty} to determine the uncertainty source, it is very hard in practice due to the difficulty of selecting an appropriate prior distribution and the expensive computation needed for Monte-Carlo estimation in RNN.

%\begin{figure}[thb]
%    % \vspace*{-1ex}
%    \begin{center}
%    \includegraphics[width=1.0\linewidth]{dirichlet.pdf}
%    \end{center}
%    \caption{Illustration of Ensembles and induced implicit distribution on the simplex $\mathbb{S}_k$.}
%    \label{fig:entropy}
%\end{figure}

\subsubsection{Prior RNN}

Inspired by the prior network in~\cite{malinin2018prior}, we propose to prior RNN explicitly parameterize model-level uncertainty at each timestamp.
Similarly, it is modeled with a distribution over distribution $p(\con|\vect{z}, \theta)$ as follows:
\begin{align*}%\label{eq:prior}
%\small
\begin{split}
    & p(\unc|\vect{z}, D)
    = \int \underbrace{p(\unc|\con) p(\con|\vect{z}, \theta)}_\text{Data} \underbrace{p(\theta|D)}_\text{Model} d\con d\theta\\
    &\approx \int p(\unc|\con) p(\con|\vect{z}, \hat{\theta}) d\con
\end{split}
\end{align*}
where 
$$p(\con| \vect{z}, \theta) = \prod_{t=1}^n p(\con_t | \con_1, \ldots, \con_{t-1}, \vect{z}; \hat{\theta})$$
$$p(\unc_t| \con) = p(\unc_t| \con_t)$$
where each $\unc_t$ is only dependent on $\con_t$.
In addition, the second step approximation is based on the following point estimate $p(\theta|D) = \delta(\theta-\hat{\theta})$ of original model uncertainty:
$$p(\con|\vect{z}, D) \approx p(\con|\vect{z}; \hat{\theta}) $$

\subsubsection{Degenerated Dirichlet Prior RNN}

% \todo{connection between $\unc$ and $\con$}

Next, we focus on how to model high-order distribution $\con_t$ at each timestamp $t$.
Following~\cite{gal2016uncertainty}, the probability density function (PDF) of Dirichlet distribution over all possible values of the $K$-dimensional categorical distribution is written as:
\begin{align*}
Dir(\con|\dir) = \begin{cases}
\frac{1}{B(\dir)}\prod_{i=1}^K \mu(i)^{\alpha(i) - 1} \quad & \text{for} \; \con \in \mathbb{S}_k\\
0 \qquad & \text{otherwise}
\end{cases}
\end{align*}
where $\dir = [\alpha(1), \cdots, \alpha(K)]^T$ with $\alpha(i) > 0$ is the concentration parameter of the Dirichlet distribution and $B(\dir)=\frac{\prod_i^K \Gamma(\alpha(i))}{\Gamma(\sum_i^k \alpha(i))}$ is the normalization factor. 
The entropy-based uncertainty measure $U(\dir)=H(Dir(\dir))$ is proven in~\cite{malinin2018prior} that can perfectly separate model-level uncertainty from data-level uncertainty and computationally efficient due to its due to its tractable statistical properties.

The Dirichlet prior RNN is practically realized by a neural network function $g$ with parameters $\theta$ to generate $n$ $K$-dimensional vectors: 
$$g(D, \theta) = (\dir_1, \ldots, \dir_n)$$
where $\dir_t \in \mathbb{R}^k$ at each timestamp $t$ follows:
$$Dir(\dir_t) = p(\con_t | \con_1, \ldots, \con_{t-1}, \vect{z}; \hat{\theta})$$
It is typically trained using both IND training data $\vect{x}$ and OOD training data $\vect{z}$.
For example, \cite{malinin2018prior} introduced a multi-task training assuming that both IND and OOD data are available (refer to the original paper for details).
However, this is not applicable in our case since OOD data is not available.
%More importantly, the defined loss functions in \cite{malinin2018prior} \todo{hard for RNN training}.

As such, we propose to degenerate Dirichlet prior RNN into the traditional RNN training only using IND data.
The posterior label distribution at each timestamp $t$ will be given by means of Dirichlet distribution:
%\begin{eqnarray*}
%	&& p(\unc_t | \unc_1, \ldots, \unc_{t-1}, \vect{z}; \hat{\theta}) \\
%%	&=& \int p(\unc_t|\con_t) p(\con_t | \con_1, \ldots, \con_{t-1}, \vect{z}; \hat{\theta})  d\con\\
%	&=& \Big[ {\alpha_t(1) \over \alpha_t(0)}, \ldots, {\alpha_t(n) \over \alpha_t(0)} \Big]
%\end{eqnarray*}
\begin{equation*}
	p(\unc_t | \unc_1, \ldots, \unc_{t-1}, \vect{z}; \hat{\theta})
	= \Big[ {\alpha_t(1) \over \alpha_t(0)}, \ldots, {\alpha_t(n) \over \alpha_t(0)} \Big]
\end{equation*}

During training, Dirichlet prior RNN $g(D, \theta)$ is optimized to maximize the empirical marginal likelihood on a training set $D = (\vect{x}, \vect{y}) \in D$:
\begin{equation*}
\begin{split}
    & \mathcal{L}_{\text{PriorRNN}} = E_{(\vect{x}, \vect{y}) \in D} [\log p(\vect{y} | \vect{x})] \\
    &= E_{(\vect{x}, \vect{y}) \in D} \Big[ {1 \over n} \sum_{j=1}^K \sum_{t=1}^n y_t(j) \log \int_{\unc_t} p(\unc_t|\vect{x})d\unc_t  \Big] \\ 
    &= E_{(\vect{x}, \vect{y}) \in D} \Big[ {1 \over n} \sum_{j=1}^K \sum_{t=1}^n y_t(j) \log \Big( \frac{\alpha_t(j)}{\alpha_t(0)} \Big) \Big]
    \label{eq:prnnloss}
\end{split}
\end{equation*}
Let $\boldsymbol{m}_t = RNN_t(\vect{x};\theta)$ be the last layer output at timestamp $t$ in RNN, it is easy to see that maximizing $\mathcal{L}_{\text{PriorRNN}}$ is equivalent to maximizing sequence cross-entropy loss in RNN if we set:
\begin{equation}
\dir_t = Me^{\boldsymbol{m}_t}= M\exp(RNN(x_t; \theta))
\end{equation}
where $M$ is the scale constant\footnote{For simplicity, we set the $M=1$ during our experiments to avoid fine-tuning the scaling hyper-parameter.}.

%Therefore, if the exponential output $e^{DNN(x; \theta)}$ of a pre-trained DNN is used as concentration parameters $\dir$ for prior network, then training softmax-based neural network is equivalent to training Dirichlet prior network. Therefore, we can easily obtain the predictive uncertainty measure as $H(Dir(\con|\exp(\boldsymbol{m})))$.

%\begin{eqnarray}
%\mathcal{L}_{RNN} = \sum_{(\vect{x}^i, \vect{y}^i) \in D} \log(\frac{\exp(m_t^i)}{\sum_i \exp(m_t^i)})
%\label{eq:ce}
%\end{eqnarray}

\emph{Remarks:
Thanks to the degeneration of Dirichlet prior RNN, the training will be the same as training of traditional RNN-based models.
That said, the Dirichlet prior RNN model is exactly the same as the selected softmax based slot filling RNN model.
This is also applicable for recent Transformer base models.
}

% Then, the predictive uncertainty measure for each word $x_t$ in an utterance $\vect{x}$ can be obtained by the entropy of learned Dirichlet  $H(Dir(\exp(\boldsymbol{m}_t)))$.

%\input{body/prior_nets}
\subsection{Dirichlet Concentration Calibration}

While the proposed degenerated Dirichlet prior RNN is sufficient in some relatively simpler case, the uncertainty measure is still sensitive and erratic to noises that its detection accuracy can hardly provide accurate estimation for model-level uncertainty.
This is mainly because that Dirichlet prior RNN is trained only on IND data.
Specifically, it is caused by the known over-fitting issue in the pre-trained neural network, where the model becomes over-confident about its prediction though it mislabels it.
In our problem of modeling uncertainty for each word in a sequence, this drawback is further amplified due to the overconfidence from both the current word and the utterance context.
Moreover, the classification model greatly emphasizes certain dimensions in the concentration parameter regardless of the form of inputs, which causes both data sources to have indistinguishable high confidence.

%\begin{figure}[thb]
%    \begin{center}
%    \includegraphics[width=0.9\columnwidth]{calibration}
%    \end{center}
%    \caption{Illustration of concentration parameters scaling $\dir$ in Prior Network for in and out-of-distribution data points.}
%    \label{fig:calibration}
%\end{figure}

Our goal is to further make the IND and OOD concepts more separable using only IND data.
Previous works~\cite{kurakin2016adversarial,liang2017enhancing} have designed various approaches to perturb the original data to simulate OOD samples.
Unfortunately, these approaches are not applicable onto natural language processing problems due to it discrete nature.

\subsubsection{Temperature Scaling Calibration}

We are inspired by ODIN~\cite{liang2017enhancing}, in which they observed the temperature scaling on last layer logits can enlarge the distance between IND and OOD sample in latent space and better separate them.
While simply using a fixed temperature $T$ to scale each parameter in $\dir_t$, i.e., $\hat{\alpha}_t(i) = \alpha_t (i)^{1/T}$, will not directly work here, we design a parameterized calibration function $\boldsymbol{\epsilon}(\dir_t; \vect{W}_c)$ with a hyperparameter $\vect{W}_c$ to learn how to scale and calibrate the concentration during model training.
At each timestamp $t$, it takes as input the concentration parameter $\dir_t$ to generate a noise in a way that it can widen the uncertainty difference between IND and OOD concepts: 
\begin{align}
   \tilde{\dir}_t = \dir_t - \boldsymbol{\epsilon}(\dir_t;\delta) = (\vect{I} - \vect{W}_c) \dir_t
\end{align}
where we apply the simplest linear transform based calibration function $\boldsymbol{\epsilon}(\dir_t;\delta) = \vect{W}_c \dir_t$  with $\vect{W}_c \in \mathbb{R}^{K*K}$ denoting the trainable calibration matrix with $K$ output labels.

\subsubsection{Joint Multi-task Training}

\begin{figure}[t]
	\centering
	\includegraphics[width=0.6\columnwidth]{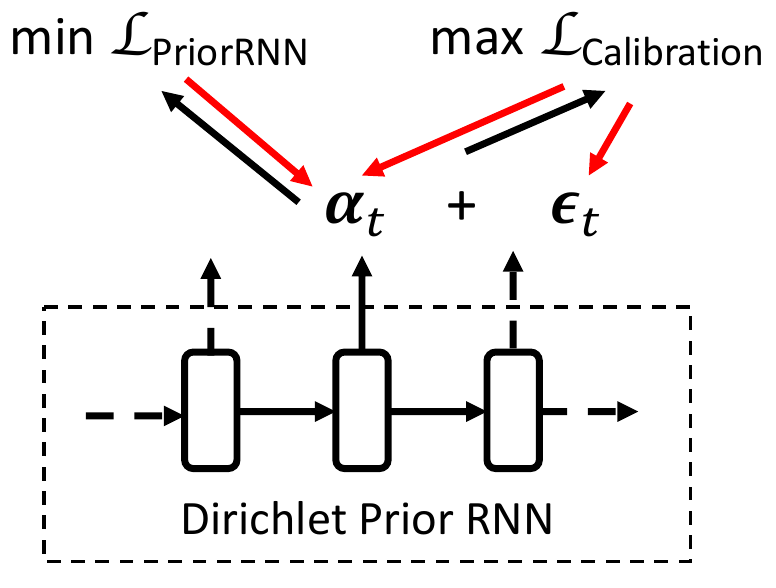}
	\caption{Multitask Training: The objective functions of both tasks are optimized simultaneously and red lines show the backpropagation.}
	\label{fig:training}
\end{figure}

~\autoref{fig:training} show the overview of our joint multi-task training.
The first task is to minimize the traditional sequence loss in~\autoref{eq:prnnloss} with the supervision of sequence label $\vect{y}$:
\begin{equation*}
\min \mathcal{L}_{\text{PriorRNN}} = \expect{(\vect{x},\vect{y}) \sim \text{IND}} \sum_{j=1}^K \sum_{t=1}^n y_t \log P_t(j)
\end{equation*}
where $P_t(j) = \Big( \frac{\exp(RNN(x_t; \theta))(j)}{\sum_{l=1}^K \exp(RNN(x_t; \theta))(l)} \Big)$.

The second task is to maximize the entropy on the calibrated Dirichlet output:
\begin{align*}\label{eq:entropyloss}
\begin{split}
& \max \mathcal{L}_{\text{Calibration}} = \expect{(\vect{x},\vect{y}) \sim \text{IND}} \Big[\sum_{t=1}^n H(\tilde{\dir_t}(x_t)) \Big] \\
& \text{s.t.} \; \forall t\;\; ||\boldsymbol{\epsilon}_t||_{\infty} < \delta ||\dir_t||_{\infty}; \;\;
\vect{W}_c \geq 0
\end{split}
\end{align*}
where $H(\cdot)$ is the entropy of Dirichlet $Dir(\tilde{\dir_t})$.
As one can see above, we consider the three constraint for this task.
First, since we do not have OOD data during training, the magnitude of such calibration matrix is encouraged to be small so that the generated noise $\boldsymbol{\epsilon}_t$ does not destroy the original IND data distribution.
Therefore we enforce the first constraint on the norm of calibrated noise: $||\boldsymbol{\epsilon}_t||_{\infty} < \delta ||\alpha_t||_{\infty}$ with $\delta$ denoting the maximum allowed calibration ratio.
Second, we enforce the non-negativity of noise $\boldsymbol{\epsilon}_t$, i.e., $\vect{W}_c \geq 0$, to ensure the negative noises.
%At last, since the calibrated concentration $\tilde{\dir}$ should still lie in the support space $\tilde{\alpha}_t(i) > 0$, we enforce the positivity of $\tilde{\dir_t} > \vect{0}$. 
The first constraint is realized by re-scaling the noise whose norm is larger than $\delta$ while the second and third constraints are realized by simply adding a ReLU~\cite{nair2010rectified} activation to $\vect{W}_c$. 

%Since it's impractical to assume we have access to out-of-distribution (OOD) dataset, we propose to use adversarial examples generated by FSGM~\cite{kurakin2016adversarial} as the synthesized OOD examples. Thus, the optimal calibration matrix $\vect{W}_c$ is described as follows:
%
%
%In order to obtain such calibration matrix $\vect{W}_c$, we propose a discriminative loss function $\mathcal{L}(\vect{W}_c)$, which aims at enlarging the gap between in- and out-of-distribution images.
%\begin{equation}
%\small
%\mathcal{L}(\vect{W}_c) = \expect{x \sim IND} [U(\tilde{\dir}(x))] - \expect{x \sim OOD} [U(\tilde{\dir}(x))]
%%    where \quad \tilde{\dir}(x) = \dir(x) + \vect{W}_c \dir(x)
%\end{equation}
%
%\begin{equation}
%\small
%    \vect{W}_c = \argmax_{\vect{W}_c} \expect{x \sim IND} [U(\tilde{\dir}(x)) - U(\tilde{\dir}(FGSM(x)))]
%%    s.t. \quad \frac{||\vect{W}_c\dir||_{\infty}}{||\dir||_{\infty}} \leq  \delta \And \vect{W}_c \geq 0 
%\end{equation}
%Here we propose to optimize $\vect{W}_c$ by gradient ascent algorithm. The first constraint is realized by re-scaling the noise whose norm is larger than $\delta$ while the second constraint is realized by simply adding a ReLU~\cite{nair2010rectified} activation to the calibration weight $\vect{W}_c$. 

% \todo{add uncertainty metrics}
% \todo{add threshold discussion connecting to table 1}

\subsection{Uncertainty Decision Engine}

Thanks to the modeling of Dirichlet distribution $\mu$, we will use its entropy $H$ to computer the uncertainty score for each word in an utterance.
For a Dirichlet distribution $Dir(\dir)$, it has a close-form entropy form as follows:
\begin{equation*}\scriptsize
\begin{split}
H(\dir) =  \log B(\dir) + (\alpha(0) - K) \psi(\alpha(0)) 
 - \sum_i^k (\alpha(i) - 1)\psi(\alpha(i))
\end{split}
\end{equation*}
where $\alpha(0)=\sum_i^k \alpha(i)$ denotes the sum over all $K$ dimensions.
Thus, the uncertainty score $u_t$ for the $t^{th}$ word in utterance $z$ can be computed as:
\begin{equation*}
u_t =
\begin{cases}
H (\dir_t) & \mbox{Dirichlet Prior RNN} \\
H (\tilde{\dir}_t) & \mbox{Calibrated Dirichlet Prior RNN} \\
\end{cases}
\end{equation*}
where Dirichlet Prior RNN has $\dir_t = Me^{\boldsymbol{m}_t}= M\exp(RNN(x_t; \theta))$ and Calibrated Dirichlet Prior RNN has calibrated $\tilde{\dir}_t = (\vect{I} - \vect{W}_c) \dir_t$.
When $u_t$ is larger than a threshold $\theta$, we set its uncertainty label as high(H) and mark it as ``unknown" by ignoring its predicted slot label.

% In (Dirichlet Prior) RNN and Calibrated Prior RNN, we use two metrics to calculate their uncertainty scores:

% \noindent\textbf{Confidence-based Uncertainty:}
% \begin{equation*}
% u_t = 1- \Big\{ \max_i {\alpha_t(i) \over \alpha_t(0)}\;\; \text{or}\;\; \max_i {\tilde{\alpha}_t(i) \over \tilde{\alpha}_t(0)} \Big\}
% %u_t = 1- \max_i \alpha_t(i) / \alpha_t(0)
% \end{equation*}

%Note that we refer to confidence measure as the negative value of entropy value $C(\dir)=-H(Dir(\con|\dir))$.

%($Dir(\tilde{\dir}_t)$) at each timestamp $t$
%Both metrics can be seen as measures of the total uncertainty (model uncertainty and data uncertainty) in predictions.

\subsection{Unknown Concept Extraction}

% \noindent\textbf{Dependency Parser (Syntax):}

We extract unknown concept by expanding the set of high uncertainty words with dependency tree of the input utterance.
% syntax with our method and some uncertainty metrics to expand the set of unknown words (Figure 2 in~\cite{jia2017concepts}).
For each word labeled as ``unknown", we traverse the dependency tree upwards along edges (including compound nouns, adjectival modifiers, or posessive modifiers) and locate the root of the corresponding noun phrase containing this word.
Then, we mark this root node and all its descendants as ``unknown".
This procedure helps to improve the recall by considering syntactically coherent phrases.

At last, we convert the labels to IOB format at phrase level by extracting all sets of consecutive words that are labeled as ``unknown".
In each set, we label the first word as ``B-unknown" and all following words as ``I-unknown".

\section{Experimental Evaluation}

Our experimental evaluation is mainly focused on the following research questions:
\begin{packed_enum}
%	\item Does Dirichlet Prior RNN model improve upon traditional RNN-based models?
	\item Can our model achieve comparable performance on IND testing set?
%	\item Does our uncertainty model deteriorate the performance of extracting IND concepts?
	\item Does entropy in degenerated Dirichlet Prior RNN better estimate uncertainty than confidence in traditional RNN models?
	\item Does concentration calibration improve upon Dirichlet Prior RNN model?
%	\item Is utterance syntax helpful?
	\item Does our method outperform existing SOTA competitors without OOD training data?
	% \item Does our method outperform the model with randomly sampled OOD training data? 
%	\item Does Dirichlet Prior RNN model help to directly label any new concepts correctly without the need of further clarification?
\end{packed_enum}

\setlength{\tabcolsep}{0.35em}
\begin{table*}[t]\scriptsize
	\centering
	\caption{Statistics of Constructed OOD Testing Set with New Concepts}
    \begin{tabular}{l|cccc|cccc}
    \toprule
    \multicolumn{1}{c|}{\multirow{2}[2]{*}{\textbf{Datasets}}} & \multicolumn{4}{c|}{\textbf{IND Training/Testing Set}} & \multicolumn{4}{c}{\textbf{OOD Testing Set with New Concepts}} \\
          & \#Concept Types & \#Train & \#Dev & \#Test & \#Utterances & \#Concepts & \#Concepts/Per Utterance & \#Words/Per Concept \\
    \midrule
    Concept Learning \cite{jia2017concepts} & 5     & 1,534 & 219   & 440   & 594   & 140   & 1.08  & 3.67 \\
    Snips \cite{Snips_dataset} & 72    & 13,084 & 700   & 700   & 4,816  & 536   & 1.21  & 3.49 \\
    ATIS \cite{Hemphill1990ATIS} & 79    & 4,478 & 500   & 893   & 1,487  & 242   & 2.76  & 1.87 \\
    \bottomrule
    \end{tabular}%
	\label{tab:statistics}%
\end{table*}%

\subsection{Datasets \& Settings}

We evaluate our method on three SLU benchmark datasets: Concept Learning~\cite{jia2017concepts}, Snips~\cite{Snips_dataset} and ATIS~\cite{Hemphill1990ATIS}.
For Snips and ATIS datasets, we collect new concepts in a subset of slot types which have many potential uncovered concepts.
The dataset details are in~\autoref{tab:statistics}.

\noindent\textbf{Concept Learning Dataset:}
is a dialogue dataset with clarification questions collected by~\cite{jia2017concepts}.
It contains 2,193 first-turn basic utterances with pre-collected concepts, and 594 first-turn utterances with unseen (personalized) concepts.
We use the original splitted 1,534 train, 659 development, and 440 test utterances for training our model.
Then we use all 594 utterances for testing our approach.
It can be downloaded from \url{http://stanford.edu/~robinjia/data/concept_learning_data.tar.gz}

\noindent\textbf{Snips Dataset:}
is a custom intent engine dataset~\cite{Snips_dataset} collected by Snips for SLU model evaluation.
It originally has 13,084 train utterances and 700 basic test (IND) utterances.
We further split all train utterances to be 12,384 train and 700 development.
It can be downloaded from \url{https://github.com/snipsco/nlu-benchmark/tree/master/2017-06-custom-intent-engines}

%It contains 7 intents.
%In each intent, the training set contains 1,800 to 2,000 utterances and the testing set contains around 100 utterances.

\noindent\textbf{ATIS Dataset:}
referred to as  \emph{Airline Travel Information Systems}, is a widely used benchmark dataset in SLU research~\cite{Hemphill1990ATIS} from the ATIS-2 and ATIS-3 corpora.
The original contains 4,478 train, 500 development and 893 basic test (IND) utterances.
It can be downloaded from \url{https://github.com/yvchen/JointSLU/tree/master/data}
%There are 79 distinct slot labels.

For Snips and ATIS datasets, we collect new concepts in a subset of slot types which have many potential uncovered concepts.
We separate the slot types to be personalized and populational.
We collect the new concepts from a group of crowd sourced workers.
For personalized concepts (e.g., playlist), we allow them to freely create any new concept.
For population concepts (e.g., movie name), we further evaluate the collected concepts by soliciting judgments from another group of crowd sourced markers.
Each turker marker is presented 50 concepts of one slot type and required to select the correct ones.
To judge the trustworthiness of each turker marker, we include 10 gold-standard concepts internally marked by experts.
Finally, we keep all population concepts that are marked correct by trustworthy markers and all collected personalized concepts.
At last, we substitute these new concepts in original basic test set (IND) and construct the new test set (OOD).
In OOD test data, we label the unseen concept as unknown in IOB format to preserve phrase level information for evaluation.
~\autoref{tab:statistics} shows the data details and statistics.

%Since each domain in Snips dataset contains completely different set of slots and very few vocabulary are shared between the domains, we treat each domain as a separate dataset.
%As such, we test both \kmodel and baseline approaches separately on each domain.
%\todo{statistics of number of words in a new concept}
%\todo{statistics of number of concepts in an utterance/slot type}

We evaluate on the state-of-the-art RNN based slot filling model~\cite{Goo2018slotgated}.
We use the state-of-the-art RNN based slot filling model \cite{Goo2018slotgated} in all of our experiments.
We use all the hyperparameters in original paper~\cite{Goo2018slotgated} for ATIS and Snips datasets.
For Concept Learning dataset, we tested different hyperparameters including hidden state size 32, 64, 128 and 256 through validation.
The same hidden state size 64 in original paper also gives the best results in Concept Learning dataset and we used it in our experiment.

For the only hyperparameter $\delta$ in our method, we experimented with different values of $\delta$ from 0.05 to 0.5 with uniform interval 0.05 (we try 10 values of $\delta$) and found that setting $\delta=0.1$ yield the best results.
We implement our method in Tensorflow using Adam optimizer with 16 mini-batch size.
On each dataset, we train both our RNN and Calibrated Prior RNN models with 20 epochs using single GPU.
All experiments are run on NVIDIA Tesla V100 GPU 16GB.

% The implementation details are in~\autoref{ap:implementation}.
% \footnote{\small We will release both code and data upon legal approval.}.

\setlength{\tabcolsep}{0.8em}
\begin{table}[t]\scriptsize
	\centering
	\caption{Slot Filling F1 Scores on Basic Test (IND) Data}
	\begin{tabular}{c|cc}
		\toprule
		\multicolumn{1}{c|}{\textbf{Dataset \textbackslash Model}} & \textbf{Calibrated Prior RNN} & \textbf{RNN Baseline} \\
		\midrule
		Concept Learning & 80.76 & 80.89 \\
		Snips & 88.56 & 88.76 \\
		ATIS  & 95.06 & 95.19 \\
		\bottomrule
	\end{tabular}%
	\label{tab:IND}%
\end{table}%

\setlength{\tabcolsep}{0.7em}
\begin{table*}[t]\scriptsize
	\centering
	\caption{\small New Concept (OOD) Extraction Results on Concept Learning Dataset without OOD Training Data (Gray: SOTA in~\cite{jia2017concepts,li2018confidence})}
	\begin{tabular}{l|c|ccc|c|ccc}
		\toprule
		\multicolumn{1}{c|}{\multirow{3}[6]{*}{\textbf{Uncertainty Metric \textbackslash Model}}} & \multicolumn{4}{c|}{\textbf{Calibrated Prior RNN} (F1 -1\% on Dev Set)} & \multicolumn{4}{c}{\textbf{RNN} (F1 -1\% on Dev Set)} \\
		\cmidrule{2-9}          & \textbf{Basic Test (IND)} & \multicolumn{3}{c|}{\textbf{New Concept Test (OOD)}} & \textbf{Basic Test (IND)} & \multicolumn{3}{c}{\textbf{New Concept Test (OOD)}} \\
		\cmidrule{2-9}          & \textbf{F1} & \textbf{Precision} & \textbf{Recall} & \textbf{F1} & \textbf{F1} & \textbf{Precision} & \textbf{Recall} & \textbf{F1} \\
		\midrule
		Dropout Perturbation & \multirow{13}[2]{*}{79.24} & -     & -     & -     & \multirow{13}[2]{*}{79.96} & \cellcolor[rgb]{ .906,  .902,  .902}25.62 & \cellcolor[rgb]{ .906,  .902,  .902}36.04 & \cellcolor[rgb]{ .906,  .902,  .902}29.95 \\
		Gaussian Noise &       & -     & -     & -     &       & \cellcolor[rgb]{ .906,  .902,  .902}24.98 & \cellcolor[rgb]{ .906,  .902,  .902}36.65 & \cellcolor[rgb]{ .906,  .902,  .902}29.71 \\
		Variance of Top Candidates &       & -     & -     & -     &       & \cellcolor[rgb]{ .906,  .902,  .902}29.66 & \cellcolor[rgb]{ .906,  .902,  .902}39.97 & \cellcolor[rgb]{ .906,  .902,  .902}34.05 \\
		OOV   &       & -     & -     & -     &       & \cellcolor[rgb]{ .906,  .902,  .902}29.35 & \cellcolor[rgb]{ .906,  .902,  .902}41.59 & \cellcolor[rgb]{ .906,  .902,  .902}34.41 \\
		Syntax+OOV &       & -     & -     & -     &       & \cellcolor[rgb]{ .906,  .902,  .902}35.12 & \cellcolor[rgb]{ .906,  .902,  .902}38.96 & \cellcolor[rgb]{ .906,  .902,  .902}36.94 \\
		Confidence &       & 45.22 & 45.47 & 45.34 &       & \cellcolor[rgb]{ .906,  .902,  .902}28.66 & \cellcolor[rgb]{ .906,  .902,  .902}36.75 & \cellcolor[rgb]{ .906,  .902,  .902}32.20 \\
		Confidence+Syntax &       & \textbf{48.09} & \textbf{48.72} & \textbf{48.40} &       & \cellcolor[rgb]{ .906,  .902,  .902}41.11 & \cellcolor[rgb]{ .906,  .902,  .902}39.62 & \cellcolor[rgb]{ .906,  .902,  .902}40.35 \\
		Confidence+OOV &       & 33.13 & 44.64 & 38.03 &       & \cellcolor[rgb]{ .906,  .902,  .902}30.50 & \cellcolor[rgb]{ .906,  .902,  .902}40.00 & \cellcolor[rgb]{ .906,  .902,  .902}34.61 \\
		Confidence+Syntax+OOV &       & 40.73 & 53.22 & 46.14 &       & \cellcolor[rgb]{ .906,  .902,  .902}36.12 & \cellcolor[rgb]{ .906,  .902,  .902}40.23 & \cellcolor[rgb]{ .906,  .902,  .902}38.06 \\
		Dirichlet Entropy &       & 46.78 & 45.19 & 45.97 &       & 40.90 & 43.39 & 42.11 \\
		Dirichlet Entropy+Syntax &       & \textbf{49.05} & \textbf{48.03} & \textbf{48.53} &       & \textbf{44.02} & \textbf{46.85} & \textbf{45.39} \\
		Dirichlet Entropy+OOV &       & 33.02 & 44.22 & 37.81 &       & 31.67 & 43.11 & 36.51 \\
		Dirichlet Entropy+Syntax+OOV &       & 40.72 & 42.87 & 41.77 &       & 39.74 & 41.97 & 40.82 \\
		\bottomrule
	\end{tabular}%
	\label{tab:concept_learning}%
\end{table*}%

\setlength{\tabcolsep}{0.7em}
\begin{table*}[t]\scriptsize
	\centering
	\caption{\small New Concept (OOD) Extraction Results on Snips Dataset without OOD Training Data (Gray: SOTA in~\cite{jia2017concepts,li2018confidence})}
	    \begin{tabular}{l|c|ccc|c|ccc}
	    \toprule
	    \multicolumn{1}{c|}{\multirow{3}[6]{*}{\textbf{Uncertainty Metric \textbackslash Model}}} & \multicolumn{4}{c|}{\textbf{Calibrated Prior RNN} (F1 -1\% on Dev Set)} & \multicolumn{4}{c}{\textbf{RNN} (F1 -1\% on Dev Set)} \\
	    \cmidrule{2-9}          & \textbf{Basic Test (IND)} & \multicolumn{3}{c|}{\textbf{New Concept Test (OOD)}} & \textbf{Basic Test (IND)} & \multicolumn{3}{c}{\textbf{New Concept Test (OOD)}} \\
	    \cmidrule{2-9}          & \textbf{F1} & \textbf{Precision} & \textbf{Recall} & \textbf{F1} & \textbf{F1} & \textbf{Precision} & \textbf{Recall} & \textbf{F1} \\
	    \midrule
	    Dropout Perturbation & \multirow{13}[2]{*}{87.67} & -     & -     & -     & \multirow{13}[2]{*}{87.81} & \cellcolor[rgb]{ .906,  .902,  .902}31.56 & \cellcolor[rgb]{ .906,  .902,  .902}60.15 & \cellcolor[rgb]{ .906,  .902,  .902}41.40 \\
	    Gaussian Noise &       & -     & -     & -     &       & \cellcolor[rgb]{ .906,  .902,  .902}30.89 & \cellcolor[rgb]{ .906,  .902,  .902}61.09 & \cellcolor[rgb]{ .906,  .902,  .902}41.03 \\
	    Variance of Top Candidates &       & -     & -     & -     &       & \cellcolor[rgb]{ .906,  .902,  .902}44.81 & \cellcolor[rgb]{ .906,  .902,  .902}64.78 & \cellcolor[rgb]{ .906,  .902,  .902}52.44 \\
	    OOV   &       & -     & -     & -     &       & \cellcolor[rgb]{ .906,  .902,  .902}33.66 & \cellcolor[rgb]{ .906,  .902,  .902}63.02 & \cellcolor[rgb]{ .906,  .902,  .902}43.88 \\
	    Syntax+OOV &       & -     & -     & -     &       & \cellcolor[rgb]{ .906,  .902,  .902}31.80 & \cellcolor[rgb]{ .906,  .902,  .902}57.23 & \cellcolor[rgb]{ .906,  .902,  .902}40.88 \\
	    Confidence &       & 46.34 & \textbf{71.26} & 56.16 &       & \cellcolor[rgb]{ .906,  .902,  .902}45.22 & \cellcolor[rgb]{ .906,  .902,  .902}62.25 & \cellcolor[rgb]{ .906,  .902,  .902}52.39 \\
	    Confidence+Syntax &       & \textbf{46.50} & 71.07 & \textbf{56.22} &       & \cellcolor[rgb]{ .906,  .902,  .902}45.27 & \cellcolor[rgb]{ .906,  .902,  .902}63.29 & \cellcolor[rgb]{ .906,  .902,  .902}52.56 \\
	    Confidence+OOV &       & 33.20 & 62.45 & 43.35 &       & \cellcolor[rgb]{ .906,  .902,  .902}32.93 & \cellcolor[rgb]{ .906,  .902,  .902}61.70 & \cellcolor[rgb]{ .906,  .902,  .902}42.94 \\
	    Confidence+Syntax+OOV &       & 30.95 & 56.29 & 39.94 &       & \cellcolor[rgb]{ .906,  .902,  .902}30.53 & \cellcolor[rgb]{ .906,  .902,  .902}69.06 & \cellcolor[rgb]{ .906,  .902,  .902}42.34 \\
	    Dirichlet Entropy &       & 47.44 & 72.26 & 57.28 &       & 46.60 & 69.25 & 55.71 \\
	    Dirichlet Entropy+Syntax &       & \textbf{47.86} & \textbf{73.51} & \textbf{57.97} &       & \textbf{46.62} & \textbf{70.33} & \textbf{56.07} \\
	    Dirichlet Entropy+OOV &       & 33.54 & 63.94 & 44.00 &       & 33.29 & 62.08 & 43.34 \\
	    Dirichlet Entropy+Syntax+OOV &       & 31.56 & 57.01 & 40.63 &       & 31.45 & 56.42 & 40.39 \\
	    \bottomrule
	    \end{tabular}%
	\label{tab:snips}%
\end{table*}%

\subsection{IND Slot Filling Results}

~\autoref{tab:IND} reports the slot filling results of IND test set.
In all datasets, we observe that our Calibrated Prior RNN model only slightly affects performance (up to 0.2 F1 drop) against RNN baseline due to the calibration loss.

\subsection{Evaluation of OOD New Concept Extraction (without OOD training data)}

We refer to our method as Dirichlet Entropy + Syntax on (Dirichlet Prior) RNN and Calibrated Dirichlet Prior RNN.

% \subsubsection{State-of-the-art (SOTA) Competitors without OOD Training Data}
\subsubsection{State-of-the-art (SOTA) Competitors}

Since existing supervised approach \cite{rao2018clarification} is not applicable in our setting,  we consider several SOTA competitors on RNN model~\cite{jia2017concepts,li2018confidence} as follows (all parameters follow original papers and the higher the more uncertain):

%\vspace{3pt}
%\noindent\textbf{RNN Confidence:}
%This metric is the same as $\max_i {\alpha_t(i) \over \alpha_t(0)}$ in Prior RNN since we degenerated it as softmax based RNN training.

\vspace{3pt}
\noindent\textbf{Dropout Perturbation:}
perform $F$ forward passes through the network, and collect the results using the perturbed parameters with dropout on a Bernoulli distribution with rate 0.25.
The variance of results are defined as uncertainty metric.

\vspace{3pt}
\noindent\textbf{Gaussian Noise:}
perform $F$ forward passes through the network, and collect the results using he perturbed parameters with Gaussian noise $N(0,0.01)$.
The variance of results are defined as uncertainty metric.

\vspace{3pt}
\noindent\textbf{Variance of Top Candidates:}
 use the variance of the probability of the top $K$ candidates ($K=5$).

\vspace{3pt}
\noindent\textbf{Out-of-Vocabulary (OOV, a.k.a. UNK):}
first constructs a vocabulary not belonging to any concept in training set:
\begin{equation*}
	\hat{V} = \{w: \exists (x_t,y_t) \in \text{IND} \; \text{s.t.} w=x_t, y_t=\text{`O'} \}
\end{equation*}
For each word $z_t$ in $\vect{z}$, we label it as ``unknown" if it is predicted as `O' but $z_t \notin \hat{V}$.

\vspace{3pt}
In addition, we also consider evaluate using \textbf{confidence} as uncertainty metrics on both $\dir_t$ in (Dirichlet Prior) RNN and $\tilde{\dir}_t$ in Calibrated Dirichlet Prior RNN:
\begin{equation*}
u_t = - \Big\{ \max_i {\alpha_t(i) \over \alpha_t(0)}\;\; \text{or}\;\; \max_i {\tilde{\alpha}_t(i) \over \tilde{\alpha}_t(0)} \Big\}
\end{equation*}
To be consistent, we consider negative confidences such that higher values indicate more uncertain.

\setlength{\tabcolsep}{0.7em}
\begin{table*}[t]\scriptsize
	\centering
	\caption{\small New Concept (OOD) Extraction Results on ATIS Dataset without OOD Training Data (Gray: SOTA in~\cite{jia2017concepts,li2018confidence})}
	    \begin{tabular}{l|c|ccc|c|ccc}
	    \toprule
	    \multicolumn{1}{c|}{\multirow{3}[6]{*}{\textbf{Uncertainty Metric \textbackslash Model}}} & \multicolumn{4}{c|}{\textbf{Calibrated Prior RNN} (F1 -1\% on Dev Set)} & \multicolumn{4}{c}{\textbf{RNN} (F1 -1\% on Dev Set)} \\
	    \cmidrule{2-9}          & \textbf{Basic Test (IND)} & \multicolumn{3}{c|}{\textbf{New Concept Test (OOD)}} & \textbf{Basic Test (IND)} & \multicolumn{3}{c}{\textbf{New Concept Test (OOD)}} \\
	    \cmidrule{2-9}          & \textbf{F1} & \textbf{Precision} & \textbf{Recall} & \textbf{F1} & \textbf{F1} & \textbf{Precision} & \textbf{Recall} & \textbf{F1} \\
	    \midrule
	    Dropout Perturbation & \multirow{13}[2]{*}{94.14} & -     & -     & -     & \multirow{13}[2]{*}{94.26} & \cellcolor[rgb]{ .906,  .902,  .902}32.23 & \cellcolor[rgb]{ .906,  .902,  .902}63.54 & \cellcolor[rgb]{ .906,  .902,  .902}42.77 \\
	    Gaussian Noise &       & -     & -     & -     &       & \cellcolor[rgb]{ .906,  .902,  .902}31.99 & \cellcolor[rgb]{ .906,  .902,  .902}64.12 & \cellcolor[rgb]{ .906,  .902,  .902}42.68 \\
	    Variance of Top Candidates &       & -     & -     & -     &       & \cellcolor[rgb]{ .906,  .902,  .902}34.61 & \cellcolor[rgb]{ .906,  .902,  .902}68.18 & \cellcolor[rgb]{ .906,  .902,  .902}45.91 \\
	    OOV   &       & -     & -     & -     &       & \cellcolor[rgb]{ .906,  .902,  .902}31.83 & \cellcolor[rgb]{ .906,  .902,  .902}68.71 & \cellcolor[rgb]{ .906,  .902,  .902}43.51 \\
	    Syntax+OOV &       & -     & -     & -     &       & \cellcolor[rgb]{ .906,  .902,  .902}31.53 & \cellcolor[rgb]{ .906,  .902,  .902}67.35 & \cellcolor[rgb]{ .906,  .902,  .902}42.95 \\
	    Confidence &       & 35.64 & \textbf{68.25} & 46.83 &       & \cellcolor[rgb]{ .906,  .902,  .902}34.27 & \cellcolor[rgb]{ .906,  .902,  .902}68.93 & \cellcolor[rgb]{ .906,  .902,  .902}45.78 \\
	    Confidence+Syntax &       & \textbf{35.73} & 68.71 & \textbf{47.01} &       & \cellcolor[rgb]{ .906,  .902,  .902}34.73 & \cellcolor[rgb]{ .906,  .902,  .902}69.21 & \cellcolor[rgb]{ .906,  .902,  .902}45.98 \\
	    Confidence+OOV &       & 31.92 & 68.48 & 43.54 &       & \cellcolor[rgb]{ .906,  .902,  .902}32.38 & \cellcolor[rgb]{ .906,  .902,  .902}69.27 & \cellcolor[rgb]{ .906,  .902,  .902}44.13 \\
	    Confidence+Syntax+OOV &       & 30.32 & 66.89 & 41.73 &       & \cellcolor[rgb]{ .906,  .902,  .902}30.74 & \cellcolor[rgb]{ .906,  .902,  .902}67.23 & \cellcolor[rgb]{ .906,  .902,  .902}42.19 \\
	    Dirichlet Entropy &       & 38.20 & 70.29 & 49.50 &       & \textbf{37.06} & 70.63 & 48.61 \\
	    Dirichlet Entropy+Syntax &       & \textbf{38.31} & \textbf{70.32} & \textbf{49.60} &       & \textbf{37.06} & \textbf{70.75} & \textbf{48.64} \\
	    Dirichlet Entropy+OOV &       & 32.51 & 69.95 & 44.39 &       & 32.96 & 70.29 & 44.88 \\
	    Dirichlet Entropy+Syntax+OOV &       & 31.03 & 68.25 & 42.66 &       & 31.50 & 68.71 & 43.20 \\
	    \bottomrule
	    \end{tabular}%
	\label{tab:atis}%
\end{table*}%

\subsubsection{Results}

\autoref{tab:concept_learning}, \ref{tab:snips} and \ref{tab:atis} report the experimental results on three different datasets.
We follow the standard metrics, precision, recall and F1 scores, in the evaluation of slot filling models.
For each dataset, we choose the confidence/entropy threshold $\theta$ at the point that F1 score on Basic Test (IND) set is reduced within 1\% against original model.
That said, our uncertainty model has slight impact on the extraction of IND concepts, which answers the first research question.
\emph{Note that for new concepts not predicted as ``unknown" but has correct slot label prediction, our experiments also consider them correct prediction.}

\vspace{3pt}
\noindent\textbf{Evaluation of Dirichlet Entropy in RNN:}
We observe that the Dirichlet entropy on (Dirichlet Prior) RNN improve the F1 score up to 5.04\% against SOTA in concept learning data with the combination of utterance syntax.
This is because the entropy with underlying high-order distribution can extract more ``unknown" words.
%Thus, even syntax does not boost (from row 3 to 4 in last column in~\autoref{tab:concept_learning}), it still leads to the significant improvement of overall performance.
Likewise, Dirichlet entropy improves F1 over SOTA by 3.51\% and 2.66\% in Snips and ATIS.
% Unfortunately, the syntax combination deteriorates the performance in these two datasets (more details in the last subsection).

\vspace{3pt}
\noindent\textbf{Evaluation of Calibrated Prior RNN:}
We observe that both confidence and entropy metrics further improve the F1 score on all three datasets.
In concept learning dataset, the F1 improvement reaches 8.18\% due to the existence of OOV words in new concepts in test set.
Such OOV words make the new concept more separable with calibrated Dirichlet concentration parameters.
In Snips and ATIS, the improvement reaches 5.41\% and 3.62\% with less OOV words in new concepts.
This follows the intuition in ODIN \cite{liang2017enhancing} for concentration calibration.

%\vspace{3pt}
%\noindent\textbf{Evaluation of OOV Words:}
%Simply labeling OOV words as ``unknown" when it is predicted as `O' leads to bad performance in all datasets.
%This is because OOV words could also exists in an utterance as non-concept words.
%
%
%\vspace{3pt}
%\noindent\textbf{Evaluation w/o Syntax:}
%It is interesting to observe that combining syntax with both our and baseline methods only improves the F1 in concept learning dataset; while even reduces the F1 in Snips  and ATIS datasets.
%This is because the new concepts in concept learning dataset are typically noun phrases, which corresponds to a subtree in dependency tree.
%In ATIS dataset, since many concepts only have a single word and the syntax parser is not perfect, the F1 score is slightly affected.
%For Snips dataset, the quite significant performance reduction with syntax is mainly because many new concepts are not in the form of noun phrase.
%For example, a book name could be a sentence and a personalized concept could be arbitrarily defined.

\vspace{3pt}
\noindent\textbf{Evaluation of SOTA competitors:}
Dropout perturbation and gaussian noise perform worst since the perturbation of model parameters break the learned distribution, especially in the context of discrete natural language data.
Variance of top candidates performs slightly better since it can partially capture the uncertainty from top candidate labels for each word.
Simply labeling OOV words as `unknown' when it is predicted as `O' leads to bad performance in all datasets since OOV could also exists as non-concept words.

It is interesting to observe that combining syntax with both our and baseline methods significantly improves the F1 in concept learning dataset and obtain slightly better F1 scores in Snips and ATIS datasets.
This is because the new concepts in concept learning dataset are typically noun phrases, which corresponds to a subtree in dependency tree.
On the other hand, many single-word concepts in ATIS and many non noun phrase concepts in Snips leads to limited help on the improvement of F1 scores.
% In Snips dataset, the quite significant performance reduction with syntax is mainly because many new concepts are not in the form of noun phrase.
% For example, a book name could be a sentence and a personalized concept could be arbitrarily defined.

In addition, we train each model 10 times with different random seeds and conduct statistical significance t-test.
As a result, our best method (Calibrated Prior RNN + Dirichlet Entropy + utterance syntax) is significantly better than all SOTA competitors (grey) on F1 score in all datasets with $\textsl{p-value}<0.01$.

\section {Conclusion}

We designed a Dirichlet prior RNN with concentration calibration for token-based uncertainty modeling to extract unknown concepts.
First, we degenerated Dirichlet into conventional RNN without changing the model training.
We further proposed a multi-task learning to calibrate the Dirichlet concentration and achieved state-of-the-art performance.
Moreover, our method can also be applied onto cutting edge Transformer based models.
In future work, we will explore the joint training with utterance syntax models.

% \todo{can be used on Transformer based models too..}

%Moreover, we will also explore to incorporate the utterance syntax naturally into joint learning of the uncertainty modeling.

%% \input{body/acknowledgement}

% \newpage

\bibliography{tagging_uncertainty,yilin,wenhu,yu,cruise,url,emnlp,iclr2019}
\bibliographystyle{acl_natbib}

% \newpage~\newpage~
% \newpage
% \appendix
% \input{appendix/dataset}
% \input{appendix/implementation}
% \input{appendix/dirichlet}

\end{document}